\documentclass[12pt,reqno,oneside]{amsart}
\usepackage[latin1]{inputenc}
\usepackage[T1]{fontenc}
\usepackage{amsmath,amsfonts,amscd,amssymb,amsthm}
\usepackage{mathtools}
\usepackage{dsfont}
\usepackage{color}
\usepackage[mathscr]{euscript}
\usepackage{comment}
\usepackage{graphicx}
\usepackage{placeins}
\usepackage{extsizes}
\usepackage{float}
\usepackage{multirow}
\usepackage{pdflscape}
\usepackage{caption}
\usepackage{subcaption}
\usepackage{tikz}
\usetikzlibrary{decorations.pathreplacing}
\definecolor{color1bg}{HTML}{f73d28}
\definecolor{color2bg}{HTML}{FA8072}
\definecolor{bblue}{HTML}{00BFFF}
\definecolor{bblue2}{HTML}{00ffff}

\usetikzlibrary{arrows,calc}
\tikzset{
>=stealth',
help lines/.style={dashed, thick},
axis/.style={<->},
important line/.style={thick},
connection/.style={thick, dotted},
}

\usetikzlibrary{shadows}
\usetikzlibrary{backgrounds}
\tikzset{
diagonal fill/.style 2 args={fill=#2, path picture={
		\fill[#1, sharp corners] (path picture bounding box.south west) -|
		(path picture bounding box.north east) -- cycle;}},
reversed diagonal fill/.style 2 args={fill=#2, path picture={
		\fill[#1, sharp corners] (path picture bounding box.north west) |- 
		(path picture bounding box.south east) -- cycle;}}
}
\usetikzlibrary{arrows.meta}

\newcommand\irregularcircle[2]{
\pgfextra {\pgfmathsetmacro\len{(#1)+rand*(#2)}}
+(0:\len pt)
\foreach \a in {10,20,...,350}{
	\pgfextra {\pgfmathsetmacro\len{(#1)+rand*(#2)}}
	-- +(\a:\len pt)
} -- cycle
}

\usepackage{rotate}

\makeatletter
\@addtoreset{equation}{section}
\makeatother
\newcounter{as}[section]

\title{The role of prior information and computational power in Machine Learning}

\author{Diego Marcondes}
\address{Department of Computer Science, Institute of Mathematics and Statistics, Universidade de S\~{a}o
	Paulo, R. do Mat\~{a}o, 1010 - Butant\~{a}, S\~{a}o Paulo - SP,
	05508-090, Brazil. \\
	e-mail: \texttt{dmarcondes@ime.usp.br}}

\author{Adilson Simonis}
\address{Department of Statistics, Institute of Mathematics and Statistics, Universidade de S\~{a}o
	Paulo, R. do Mat\~{a}o, 1010 - Butant\~{a}, S\~{a}o Paulo - SP,
	05508-090, Brazil.}

\author{Junior Barrera}
\address{Department of Computer Science, Institute of Mathematics and Statistics, Universidade de S\~{a}o
	Paulo, R. do Mat\~{a}o, 1010 - Butant\~{a}, S\~{a}o Paulo - SP,
	05508-090, Brazil.}

\allowdisplaybreaks

\usepackage{lineno}

\begin{document}
	\maketitle
	
\begin{abstract}
	Science consists on conceiving hypotheses, confronting them with empirical evidence, and keeping only hypotheses which have not yet been falsified. Under deductive reasoning they are conceived in view of a theory and confronted with empirical evidence in an attempt to falsify it, and under inductive reasoning they are conceived based on observation, confronted with empirical evidence and a theory is established based on the not falsified hypotheses. When the hypotheses testing can be performed with quantitative data, the confrontation can be achieved with Machine Learning methods, whose quality is highly dependent on the hypotheses' complexity, hence on the proper insertion of prior information into the set of hypotheses seeking to decrease its complexity without loosing \textit{good} hypotheses. However, Machine Learning tools have been applied under the pragmatic view of instrumentalism, which is concerned only with the performance of the methods and not with the understanding of their behavior, leading to methods which are not fully understood. In this context, we discuss how prior information and computational power can be employed to solve a learning problem, but while prior information and a careful design of the hypotheses space has as advantage the interpretability of the results, employing high computational power has the advantage of a higher performance. We discuss why learning methods which combine both should work better from an understanding and performance perspective, arguing in favor of basic theoretical research on Machine Learning, in special about how properties of classifiers may be identified in parameters of modern learning models.
	
\end{abstract}

\section{Introduction}

Modern science consists on conceiving a set of hypotheses to explain a phenomenon, confronting them with empirical evidence, and keeping as possible explanations only hypotheses which have not yet been falsified. 

Under deductive reasoning, the hypotheses are conceived in view of a theory and confronted with empirical evidence in an attempt to falsify the established theory and, depending on which hypotheses are not falsified, the theory stands or should be redesigned, but, either way, it will again be put to test by trying to falsify remaining hypotheses which are a consequence of it \cite{popper2005logic}. Under inductive reasoning, the hypotheses are conceived based on observation, confronted with empirical evidence via experimentation and then a theory is established based on the not falsified hypotheses, and the theory advances as more hypotheses are tested and not falsified, so the hypotheses compatible with quantitative or qualitative empirical data are generalized to the phenomena which generated the data \cite{holland1989induction}.

When the phenomena which the hypotheses pertain to are characterized by quantitative empirical evidence, the confrontation of hypotheses may be performed with Machine Learning tools, which consider as not falsified hypotheses those that well fit empirical quantitative data. The theoretical basis of such hypotheses testing is provided by Statistical Learning Theory \cite{vapnik1998}, a framework for Machine Learning concerned with inferring patterns from data and quantifying uncertainty about these inferences. Statistical Learning is mainly a tool for inductive inference \cite{angluin1983inductive}, in which patterns observed on a dataset are generalized to data generated by a same process, but not yet observed.

The generalization quality of Machine Learning methods is highly dependent on the set of hypotheses, called model or hypotheses space, and on its complexity. Highly complex models may lead to unfalsifiability when for whatever data generated by the process there is a hypothesis that fully explain it, and hence the model cannot be falsified \cite[Section~3.11]{vapnik1998}. Nevertheless, there are mathematical results about the complexity of a hypotheses space that guarantees the falsifiability of it if the sample size is great enough, what implies that the hypotheses space should be of limited complexity to apply Machine Learning methods within the scientific method.

Each hypotheses space has a complexity measure called Vapnik-Chervonenkis (VC) dimension and the Fundamental Theorem of Statistical Learning states that a hypotheses space is falsifiable for a sample great enough if, and only if, it has finite VC dimension \cite[Theorem~6.7]{shalev2014}. Furthermore, it implies that as great the sample size and as simple the hypotheses space (small VC dimension), greater is the generalization quality of the learned hypothesis, i.e., not falsified hypothesis, relative to the best generalization that can be achieved within the fixed set of hypotheses. See \cite{vapnik2000} for an overview of Statistical Learning Theory. 

This result brings a trade-off between the complexity of a model and the generalization quality of its \textit{best} hypothesis, which we define as the generalization quality of the model. On the one hand, if the model is simple, then the generalization quality of the learned hypothesis should be as good as that of the model by the Fundamental Theorem of Statistical Learning, but the model may not generalize very well since it is simple and may not be able to capture all the nuances of the phenomena being modeled. 

On the other hand, if the model is complex, then the generalization quality of the learned hypothesis should not be as good as that of the model if the sample size is not great enough, even though the model generalize well. Since the complexity is related to the generalization quality of the learned hypothesis relative to the model, which we define as the internal generalization of the learned hypothesis, this trade-off can also been seen as between the internal generalization of the learned hypothesis and the generalization of the model.

The unfalsifiability of models with infinite VC dimension, and the trade-off between the internal generalization of the learned hypothesis and the generalization of the model, theoretically demand a careful design of the hypotheses space to learn a hypothesis that well generalizes. Fortunately, this design is an integral part of both deductive and inductive reasoning, which have in common the full knowledge and control of the hypotheses space, since it has been delimited by the researcher, either based on a theory (rationalism/critical rationalism \cite{popper2014conjectures,Cottingham1988-COTTR}) or on empirical observation (empiricism \cite{Feyerabend1965-FEYPOE}), respectively.

Notwithstanding the theoretical and philosophical demand for a conscious delimitation of the hypotheses space, with the advent of high computational power, Machine Learning tools have been applied under the pragmatic methodological view of instrumentalism \cite{vapnik2006realism,jones2018we}, which is concerned solely with the performance of the learned hypothesis on empirical data and not with the understanding of its implications within a theory. This point of view led to modern Machine Learning methods which are not fully understood and comprise so-called black box approaches, but that obtain remarkable practical results and serve as instruments to perform tasks within an artificial intelligence system \cite{ghahramani2015probabilistic,michalski2013machine}. The main modern Machine Learning tool within this paradigm are neural networks \cite{charu2018neural}, which have revolutionized the application of Machine Learning in the last decade by having outstanding performance on learning tasks in many domains \cite{abiodun2018state}.

The lack of a reasoning to establish consciously and a priori the set of hypotheses led to a detachment of the modern methods from classical Statistical Learning Theory and the remarkable results observed in practice with a given sample size are not explained by the theory since, in principle, the space of the hypotheses representable by modern methods is too complex to obtain high generalization quality with the available sample size. Hence, not only the interpretability of the methods in view of a theory is lost, but there is also a lack of a mathematical basis to explain its performance. The lack of interpretability and mathematical basis led to the development of new methods on a trial and error basis based almost entirely on empirical performance, raising a discussion about the scientific nature of artificial intelligence research \cite{hutson2018has}.

The widespread use of black box instruments have also raised ethical, moral and legal issues related to data protection, privacy, potential bias and lack of transparency, explainability and accountability, among others, and have greatly affected many areas of society \cite{gill2020prediction,o2016weapons}. Even though the discussion about these issues predates the era of high computational power \cite{weizenbaum1976computer}, they are ever more prevalent in an increasing high-tech culture. The domains in which these issues arise are such that high performance is not sufficient and a better understanding and control over the task being performed are necessary, since generalization quality is not enough in domains such as education, medicine, healthcare, judiciary system and public policy, that require a reasoning to develop a theory to explain the machine decision.

In Statistical Learning Theory, the generalization quality is associated with controlling the trade-off between the internal generalization of the learned hypothesis and the generalization of the model. This is accomplished by fixing a simple model with hypotheses that ought to well generalize, and for this the researcher relies heavily on prior information about the phenomena being modeled, which can be either theoretically established (rationalism) or obtained from empirical evidence (empiricism). 

The available prior information should be translated into hypotheses; if prior information is right, the model should well generalize; and if the model is simple, high internal generalization may be achieved. In this framework prior information is vital to delimit the set of hypotheses in view of the Fundamental Theorem of Statistical Learning and this prior delimitation has the advantage of allowing the testing (deductive reasoning) or development (inductive reasoning) of a theory that explains a phenomenon. Therefore, classical methods have the tools to provide both performance and interpretability.

Since modern learning methods are detached from the classical theory and are black box instruments, they do not benefit as much from prior information as they do from high computational power. Indeed, the increase in performance of these methods over the years is mainly due to advances in hardware and computational power rather than great theoretical advances (see \cite{pandey2022transformational}). As prior information is vital to classical methods of inductive inference, computer power is vital for modern Machine Learning methods. However, although modern methods may provide a much higher performance than classical methods, they lose in interpretability, leading to issues when interpretability is strictly needed.

In this context, we discuss the role of prior information and computational power in Machine Learning methods. We discuss how both can be employed to solve a learning problem, but while prior information and a careful design of the hypotheses space has as advantage the interpretability of the results, employing high computational power has the advantage of a higher performance. We argue that learning methods which combine prior information with high computational power should work better from both the understanding and performance perspective. We present two general frameworks under which prior information and computational power may be combined to better learn. With this discussion, we seek to argue in favor of basic theoretical research about modern Machine Learning methods, such as neural networks, from a Statistical Learning perspective. In special, we make a point in favor of better understanding how properties of classifiers may be identified in parameters of modern learning models.

In Section \ref{SecML} we present an overview of Machine Learning theory from a Statistical Learning perspective. In Sections \ref{SecPrior} and \ref{SecHigh} we discuss, respectively, the role of prior information and high computational power in Machine Learning. In Section \ref{Sec_comb} we proposed two frameworks under which prior information could be combined with high computational power in order to take advantage of both approaches to have a greater understanding and performance. In Section \ref{SecFR} we present our final remarks.

\section{Machine Learning Theory}
\label{SecML}

The theoretical foundations of learning theory were established by Vapnik and Chervonenkis in the 1970s \cite{vapnik1971uniform,vapnik1974method,vapnik1974method2,vapnik1974theory} and culminated in what is now known as Vapnik-Chervonenkis (VC) theory or Statistical Learning Theory \cite{vapnik1998,vapnik2000}. VC theory is concerned with the generalization quality of a learned hypothesis, that is, how well it performs on instances that were not in the sample employed to learn it. Among learning tasks covered by VC theory, there are classification tasks, classically known as pattern recognition \cite{devroye1996}, on which we will mainly focus the discussion of this paper in order to facilitate the understanding of the ideas discussed here.

A classification task consists in classifying instances $X$, that take values in a set $\mathcal{X}$, into one of two categories $\{0,1\}$. In order to learn a classifier to perform such a task, one has a sample $\mathcal{D}_{N} = \{(X_{1},Y_{1}),\dots,(X_{N},Y_{N})\}$ of $N$ instances $X_{i}$ and their label $Y_{i}$. The instances $X_{i}$ are generated by a statistical distribution $P(x)$ and their label is sampled from a conditional distribution $P(y|x)$. Although a sample from these statistical distributions is available, they are both unknown. Learning in this context means obtaining a classifier $\hat{h}: \mathcal{X} \to \{0,1\}$ from a sample $\mathcal{D}_{N}$ which \textit{well-generalizes}, that is, works well when classifying new instances $X$ generated by $P(x)$ into labels generated by $P(y|x)$.

The first step of learning is fixing the space of all classifiers one is willing to consider as possible rules associating the instances $X$ to their label $Y$. We denote this space by $\mathcal{H} = \{h: \mathcal{X} \to \{0,1\}\} \subset \mathcal{F}$, which is a subset of the space of all functions from $\mathcal{X}$ to $\{0,1\}$ that we denote by $\mathcal{F}$. We call an element $h$ of $\mathcal{H}$ \textit{hypothesis}, and $\mathcal{H}$ is called \textit{hypotheses space} or model. Each hypothesis $h$ has an expected risk when classifying instances $X$ into labels $Y$ denoted by
\begin{linenomath}
	\begin{equation*}
		L(h) \coloneqq P(h(X) \neq Y)
	\end{equation*}
\end{linenomath}
that is the probability of $h(X)$ being different of $Y$, when $X$ is sampled from $P(x)$ and $Y$ from $P(y|x)$.

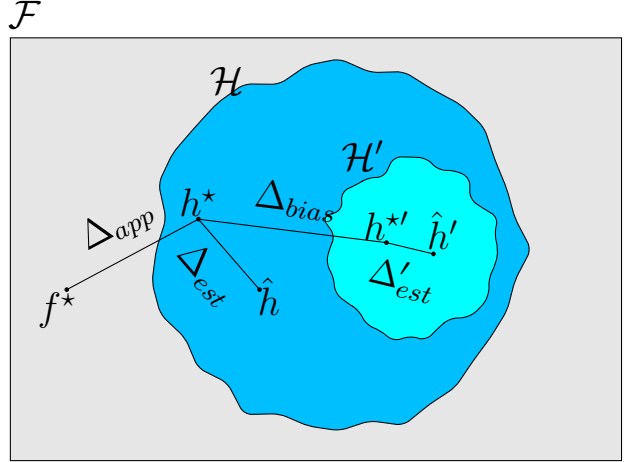
\begin{figure*}[ht]
	\begin{center}
		\begin{tikzpicture}[scale=1.25, transform shape]
			\tikzstyle{hs} = [rectangle,draw=black,fill = color2bg,rounded corners, minimum height=3em, minimum width=3.5em, node distance=2.5cm, line width=1pt]
			\tikzstyle{ns} = [rectangle,draw=white,rounded corners, minimum height=3em, minimum width=3.5em, node distance=2.5cm, line width=1pt]
			
			\draw[fill = gray!20] (0,0.5) rectangle (6.5,5);
			\draw[fill = bblue,rounded corners=1mm] (3.5,2.75) \irregularcircle{1.95cm}{1mm};
			\draw[fill = bblue2,rounded corners=1mm] (4.25,2.75) \irregularcircle{0.9cm}{1mm};
			
			\node at (2.3,4.55) {$\mathcal{H}$};
			\node at (0.15,5.25) {$\mathcal{F}$};
			\node at (3.75,3.75) {$\mathcal{H}^\prime$};
			
			\node at (2,3.25) {$h^{\star}$};
			\node at (2,3.07)[circle,fill,inner sep=0.5pt]{};
			\node at (2.75,2.25) {$\hat{h}$};
			\node at (2.65,2.32)[circle,fill,inner sep=0.5pt]{};
			\node at (0.5,2.1) {$f^{\star}$};
			\node at (0.6,2.32)[circle,fill,inner sep=0.5pt]{};
			\node at (4,3) {$h^{\star\prime}$};
			\node at (4,2.82)[circle,fill,inner sep=0.5pt]{};
			\node at (4.6,2.95) {$\hat{h}^{\prime}$};
			\node at (4.5,2.7)[circle,fill,inner sep=0.5pt]{};
			
			\draw[-] (2,3.07) -- (2.65,2.32) node[pos=0.5,sloped,below] {$\Delta_{est}$};
			\draw[-] (2,3.07) -- (0.6,2.32) node[pos=0.5,sloped,above] {$\Delta_{app}$};
			\draw[-] (4,2.82) -- (2,3.07) node[pos=0.5,sloped,above] {$\Delta_{bias}$};
			\draw[-] (4,2.82) -- (4.5,2.7) node[pos=0.5,sloped,below] {\small$\Delta^{\prime}_{est}$};;
		\end{tikzpicture}
	\end{center}
	\caption{The set $\mathcal{F}$ of all functions from $\mathcal{X}$ to $\{0,1\}$ and two subsets $\mathcal{H}, \mathcal{H}^{\prime}$ of it that are possible hypotheses spaces. The complexity of the subsets is proportional to their size. Each subset has a target hypothesis ($h^{\star}$ and $h^{\star\prime}$) and both differ from the best hypothesis $f^{\star}$ of $\mathcal{F}$. When learning on $\mathcal{H}$ and $\mathcal{H}^{\prime}$ one learns hypotheses $\hat{h}$ and $\hat{h}^{\prime}$, respectively. The approximation error $\Delta_{app}$ of $\mathcal{H}$, the estimation errors $\Delta_{est}$ and $\Delta^{\prime}_{est}$ of $\mathcal{H}$ and $\mathcal{H}^{\prime}$, respectively, and the bias $\Delta_{bias}$ of $\mathcal{H}^{\prime}$ relative to $\mathcal{H}$ are depicted.} \label{figHS}
\end{figure*}

Among the hypotheses in $\mathcal{H}$ there are those with minimum expected risk, which are hypotheses $h^{\star}$ satisfying
\begin{linenomath}
	\begin{equation*}
		L(h^{\star}) \leq L(h), \ \ \forall h \in \mathcal{H}.
	\end{equation*}
\end{linenomath}
We call $h^{\star}$ a target hypothesis of $\mathcal{H}$. It may happen that the function $f^{\star}$ which minimizes the risk in $\mathcal{F}$ is not in $\mathcal{H}$ so that
\begin{linenomath}
	\begin{equation*}
		L(f^{\star}) < L(h^{\star}),
	\end{equation*}
\end{linenomath}
but, once $\mathcal{H}$ is fixed, the learning process will consider only hypotheses in $\mathcal{H}$, hence one cannot do any better than $h^{\star}$. Therefore, there may be an intrinsic approximation error in the learning process that we denote by
\begin{linenomath}
	\begin{equation*}
		\Delta_{app} \coloneqq L(h^{\star}) - L(f^{\star}).
	\end{equation*}
\end{linenomath}
This error is illustrated in Figure \ref{figHS}.

With this notation, a learned hypothesis $\hat{h}$ well-generalizes in $\mathcal{H}$ if
\begin{linenomath}
	\begin{equation}
		\label{est_error}
		\Delta_{est} \coloneqq L(\hat{h}) - L(h^{\star}) < \epsilon
	\end{equation}
\end{linenomath}
for a $\epsilon > 0$ small enough, in which $L(\hat{h})$ is called the generalization error of $\hat{h}$. We call the quantity on the left-hand side of inequality \eqref{est_error} estimation error, which is also depicted in Figure \ref{figHS}. An important thing to note here is that the generalization quality of $\hat{h}$ is relative to that of $h^{\star}$, since when considering only hypotheses in $\mathcal{H}$ one cannot do better than $h^{\star}$. Hence, in order to open the possibility of having good generalization in an absolute sense\footnote{With absolute sense we mean a good generalization relative to $f^{\star}$, that is the best generalization possible under distribution $P$, even if $L(f^{\star}) \gg 0$.} one has to design $\mathcal{H}$ in a way such that $L(h^{\star}) \approx L(f^{\star})$, for otherwise the generalization quality of $\hat{h}$ will be limited by that of $h^{\star}$, which may not be the best that can be achieved. 

The classical solution to the learning problem is to take $\hat{h}$ as a hypothesis in $\mathcal{H}$ which minimizes the empirical risk in sample $\mathcal{D}_{N}$ given by
\begin{linenomath}
	\begin{equation*}
		L_{\mathcal{D}_{N}}(h) = \frac{1}{N} \sum_{i = 1}^{N} \mathds{1}\{h(X_{i}) \neq Y_{i}\}, \ \ h \in \mathcal{H},
	\end{equation*}
\end{linenomath}
that is the proportion of points in the sample miss-classified by $h$, in which 
\begin{linenomath}
	\begin{equation*}
		\mathds{1}\{h(X_{i}) \neq Y_{i}\} = \begin{cases}
			1, & \text{ if } h(X_{i}) \neq Y_{i}\\
			0, & \text{ if } h(X_{i}) = Y_{i}
		\end{cases}.
	\end{equation*}
\end{linenomath}
In this approach, called empirical risk minimization, $\hat{h}$ is such that
\begin{linenomath}
	\begin{equation*}
		L_{\mathcal{D}_{N}}(\hat{h}) \leq L_{\mathcal{D}_{N}}(h), \ \ \forall h \in \mathcal{H}.
	\end{equation*}
\end{linenomath}

The main result of VC theory are bounds $\Omega(\epsilon,N,d_{VC}(\mathcal{H}))$, depending on the sample size $N$ and on a measure of the complexity of $\mathcal{H}$ called VC dimension, for the probability of the expected risk $L(\hat{h})$ of the estimated hypothesis differing more than $\epsilon$ from the risk of a target hypothesis of $\mathcal{H}$, that is
\begin{linenomath}
	\begin{equation}
		\label{bound}
		P\left(L(\hat{h}) - L(h^{\star}) > \epsilon\right) \leq \Omega(\epsilon,N,d_{VC}(\mathcal{H})).
	\end{equation}
\end{linenomath}
We refer to \cite{vapnik1998,vapnik2000,devroye1996,shalev2014} for the definition of VC dimension and an explicit form of $\Omega(\epsilon,N,d_{VC}(\mathcal{H}))$. The bound $\Omega(\epsilon,N,d_{VC}(\mathcal{H}))$ is decreasing in $\epsilon$, decreasing in the sample size $N$ and increasing in the complexity of $\mathcal{H}$, measured by the VC dimension of $\mathcal{H}$ denoted by $d_{VC}(\mathcal{H})$. This implies that, for a fixed $\epsilon > 0$, bound \eqref{bound} can be tightened by either increasing the sample size $N$ or decreasing the complexity of $\mathcal{H}$ ($d_{VC}(\mathcal{H})$). 

Furthermore, for any $\epsilon$ fixed, the bound tends to zero when $N$ increases if the VC dimension is finite, meaning that hypotheses spaces with finite complexity are such that, as the sample size increases, greater is the probability of having a good generalization relative to $h^{\star}$. Actually, the probability in \eqref{bound} tends to zero if, and only if, $d_{VC}(\mathcal{H})$ is finite, that is a consequence of the Fundamental Theorem of Statistical Learning \cite[Theorem~6.7]{shalev2014}.

This implies that one has to choose a hypotheses space $\mathcal{H}$ with finite VC dimension, for otherwise not even an infinite number of samples suffice to have good generalization. Hence, when $\mathcal{F}$, the set of all functions from $\mathcal{H}$ to $\{0,1\}$, has infinite VC dimension, as is often the case, it is strictly necessary to choose a subset of it to learn on. 

From VC theory follows that the generalization quality of a learned hypothesis $\hat{h}$ (a) is limited by the risk of $h^{\star}$ and (b) for a fixed sample size $N$ is as good as that of $h^{\star}$ with higher probability as less complex $\mathcal{H}$ is. Facts (a) and (b) brings upon a trade-off between the complexity of $\mathcal{H}$, which by \eqref{bound} is related to the internal generalization error of learning on $\mathcal{H}$, and the risk $L(h^{\star})$ of a target hypothesis.

On the one hand, by increasing the complexity of $\mathcal{H}$, the risk of its target hypothesis will decrease, and may be quite low when $\mathcal{H}$ is complex enough to properly express the relation between instances $X$ and their label $Y$, but $\hat{h}$ may not generalize well since, with high probability, it may have an expected risk too greater than that of $h^{\star}$. On the other hand, if $\mathcal{H}$ has low complexity, than $L(\hat{h})$ may be close to $L(h^{\star})$ with high probability, but $L(h^{\star})$ may be too great since the hypotheses in $\mathcal{H}$ may not be able to capture all the nuances of the relation between $X$ and its label.

The generalization-approximation error trade-off, or analogously complexity-approximation error trade-off, is illustrated in Figure \ref{figHS}, which represents the space $\mathcal{F}$ of all functions from $\mathcal{X}$ to $\{0,1\}$, and two options of hypotheses spaces $\mathcal{H}$ and $\mathcal{H}^{\prime}$, whose complexities are proportional to their size. On the one hand, regarding the approximation error side of the trade-off, if one considered $\mathcal{H}$ he would have an approximation error $\Delta_{app}$, while if he considered $\mathcal{H}^{\prime}$ he would have the approximation error $\Delta_{app} + \Delta_{bias}$ in which
\begin{linenomath}
	\begin{equation*}
		\Delta_{bias} \coloneqq L(h^{\star\prime}) - L(h^{\star})
	\end{equation*}
\end{linenomath}
is the difference between the expected risk of a target hypothesis $h^{\star\prime}$ of $\mathcal{H}^{\prime}$ and that of a target of $\mathcal{H}$. We conclude that the approximation error is greater in $\mathcal{H}^{\prime}$.

On the other hand, in what concerns the internal generalization error side of the trade-off, if one considered $\mathcal{H}^{\prime}$ he would better estimate the target $h^{\star\prime}$, hence have a better generalization relative to the best hypothesis in $\mathcal{H}^{\prime}$, than if he considered $\mathcal{H}$, where he could not have a generalization as good with respect to $h^{\star}$ since $\mathcal{H}$ is more complex than $\mathcal{H}^{\prime}$. Hence, from this point of view, the internal generalization error is lower in $\mathcal{H}^{\prime}$, even though the approximation error is greater in it.

Nevertheless, adding a bias to learn on $\mathcal{H}^{\prime}$ instead of on $\mathcal{H}$ may be worth it. Indeed, when learning on $\mathcal{H}$ one commits the following error with respect to $f^{\star}$
\begin{equation*}
	L(\hat{h}) - L(f^{\star}) = \Delta_{app} + \Delta_{est},
\end{equation*}
while when learning on $\mathcal{H}^{\prime}$ he commits the error
\begin{equation*}
	L(\hat{h}^{\prime}) - L(f^{\star}) = \Delta_{app} + \Delta_{bias} + \Delta^{\prime}_{est}.
\end{equation*}
The estimation error compensates the bias of learning on $\mathcal{H}^{\prime}$ when
\begin{equation}
	\label{inequality_MS}
	\Delta_{bias} + \Delta^{\prime}_{est} <  \Delta_{est},
\end{equation}
since in this case, the generalization error is lesser on $\mathcal{H}^{\prime}$. Inequality \eqref{inequality_MS} is a theoretical guide on whether to learn on $\mathcal{H}$ or in a subset of it, and is the essence of the generalization-approximation error trade-off.

When learning one should mind both the approximation error and the complexity of the hypotheses space, balancing the trade-off, to better learn with a sample of fixed size, and compensate the possible lack of data by controlling the approximation error and the complexity of the hypotheses space. The discussion in the following sections will revolve around the control of this trade-off by incorporating prior information about the problem at hand into the learning process or employing high computational power to learn, which are actually manners of mitigating the lack of data, since if there were infinite data then \eqref{bound} would be zero for any $\mathcal{H}$ with finite VC dimension and there would only remain the approximation error side of the trade-off to be controlled. We will argue that it is possible to absolutely increase the generalization quality of a learned hypothesis $\hat{h}$,for a fixed sample size, with both approaches, but a better way of doing so from a scientific perspective is combining prior information with high computational power.

\section{Inserting prior information into Machine Learning}
\label{SecPrior}

In the absence of great datasets and high computational power, many Machine Learning applications in the decades leading up to the 2010s have relied heavily on constraining as much as possible the hypotheses space via the insertion of prior information about the problem at hand to properly learn a hypothesis with the resources available. 

An extreme example of incorporating prior information in a learning task was the early decades of an area called Mathematical Morphology, which arose in the 1960s concerned with defining and developing properties of operators designed to transform binary images \cite{serra1982image}. At first, these operators were completely hand-made and the researcher had to, based on trial and error, and its knowledge about the image domain and desired output, define the operations that had to be performed in an image to obtain a desired result. There was a toolbox \cite{barrera1994mathematical,barrera1998mmach} of building blocks which could be combined to obtain certain results in an image, and the researchers were specialized in designing by hand the operators.

\begin{figure}[ht]
	\begin{minipage}{0.5\linewidth}
		\begin{center}
			\includegraphics[width=0.8\linewidth]{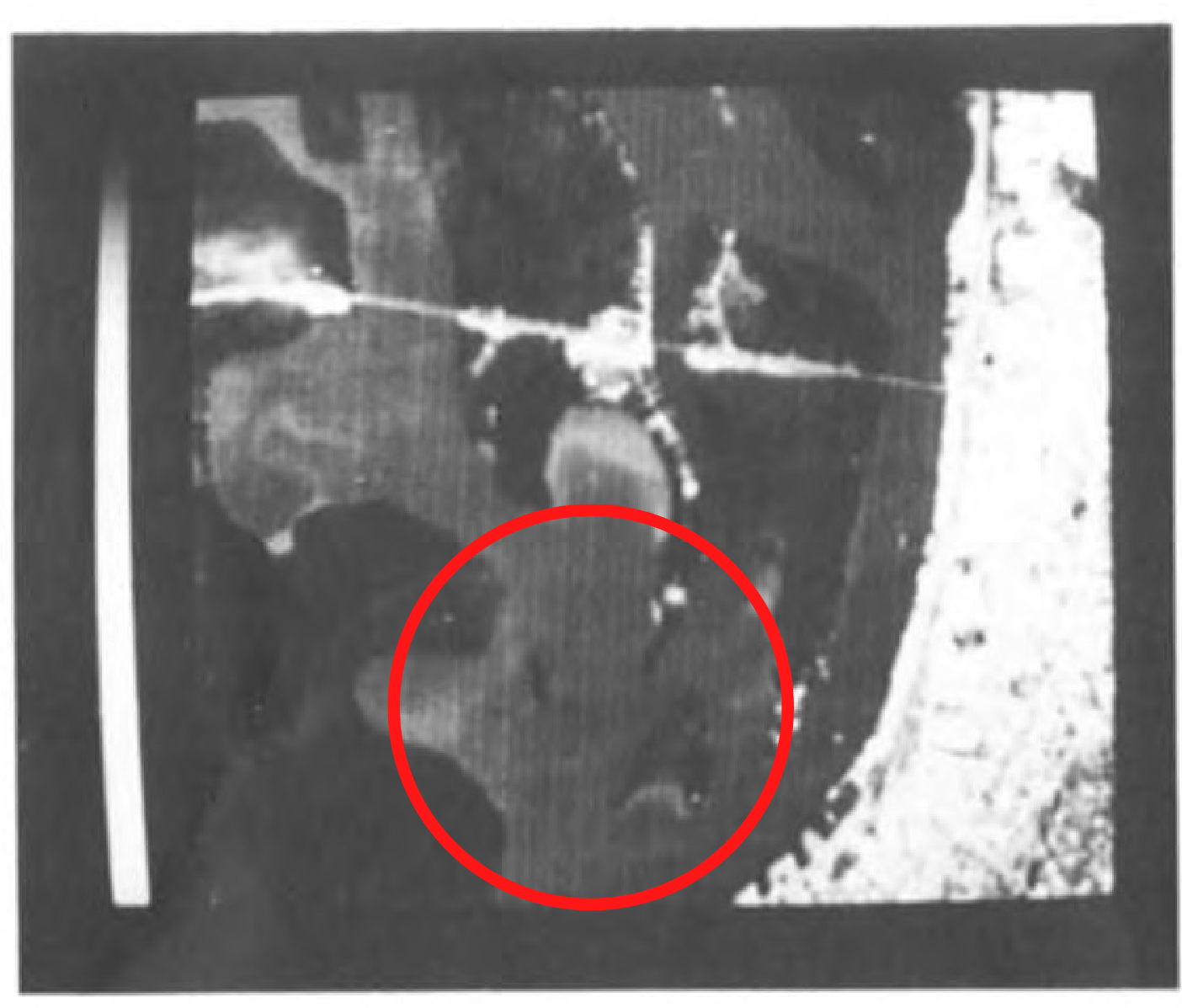}
			
			(a)
		\end{center}
	\end{minipage}
	\begin{minipage}{0.5\linewidth}
		\begin{center}
			\includegraphics[width=0.8\linewidth]{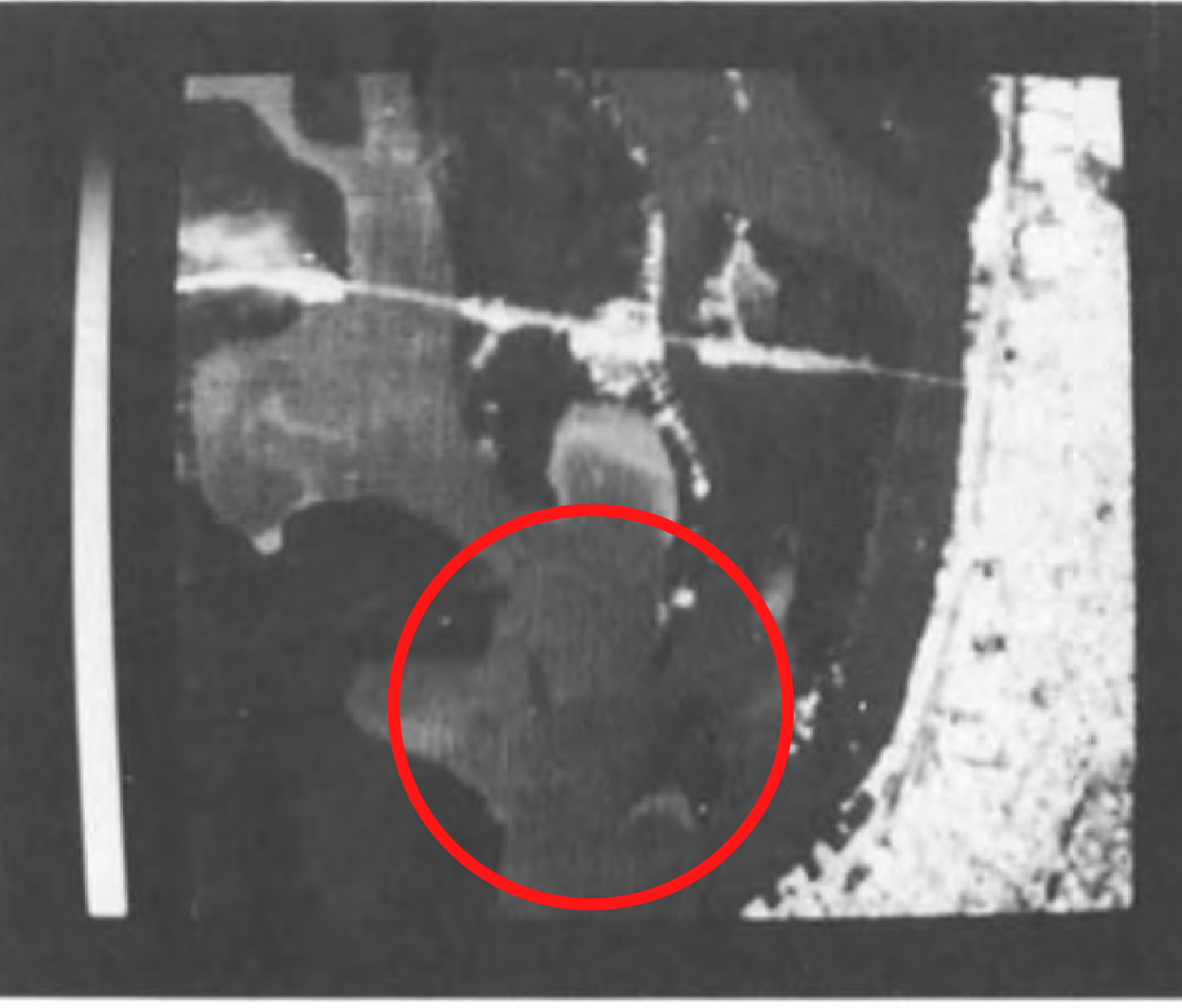}
			
			(b)
		\end{center}	
	\end{minipage}
	\caption{(a) Gray-scale original image of Porto Alegre City in Brazil obtained with the SPOT-1 satellite around 1988. The red circle outlines stripping in the image. (b) Original image after stripping correction via a hand-made morphological filter. Observe the almost absence of stripping within the red circle. These images were adapted from \cite{banon1989morphological} where they originally appeared.} \label{satalite}
\end{figure}

An example of this hand-made approach is the work of \cite{banon1989morphological}, where the authors employed prior modeling of the stripping phenomenon in satellite images to design a morphological operator for stripping correction. They accomplished the result presented in Figure \ref{satalite}, which was obtained without using any automatic method for learning an operator and only knowledge about the problem domain and properties of morphological operators were employed (see \cite{banon1989morphological} for the reasoning which led to the result in Figure \ref{satalite}). 

The example above is the extreme of incorporating prior information, since once it is performed, there is nothing to be learned as the hypotheses space contains only one hand-made hypothesis: $\mathcal{H} = \{h^{\star}\}$. Nevertheless, the usual incorporation of prior information in a learning problem happens in a step before the Statistical Learning, when the hypotheses space is carefully chosen based on prior information and then a hypothesis is learned on it from data. 

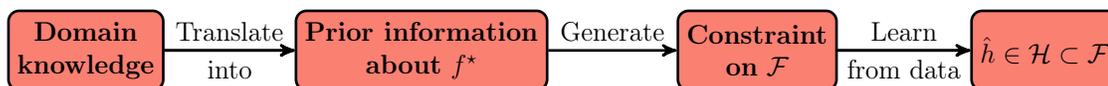
\begin{figure*}[ht]
	\begin{center}
		\begin{tikzpicture}[scale=0.85, transform shape]
			\tikzstyle{hs} = [rectangle,draw=black,fill = color2bg,rounded corners, minimum height=3em, minimum width=3.5em, node distance=2.5cm, line width=1pt,align=center]
			\tikzstyle{ns} = [rectangle,draw=white,rounded corners, minimum height=3em, minimum width=3.5em, node distance=2.5cm, line width=1pt]
			
			\node[hs] (a1) at (-11,0) {\textbf{Domain}\\ \textbf{knowledge}};
			\node[hs] (a2) at (-5.75,0) {\textbf{Prior information}\\ \textbf{about} $f^{\star}$};
			\node[hs] (a3) at (-0.5,0) {\textbf{Constraint}\\ \textbf{on} $\mathcal{F}$};
			\node[hs] (a4) at (4,0) {$\hat{h} \in \mathcal{H} \subset \mathcal{F}$};
			
			\begin{scope}[line width=1pt]
				\draw[->] (a1) -- (a2) node[pos=0.5,sloped,above] {Translate};
				\draw[->] (a1) -- (a2) node[pos=0.5,sloped,below] {into};
				\draw[->] (a2) -- (a3) node[pos=0.5,sloped,above] {Generate};
				\draw[->] (a3) -- (a4) node[pos=0.5,sloped,above] {Learn};
				\draw[->] (a3) -- (a4) node[pos=0.5,sloped,below] {from data};
			\end{scope}
		\end{tikzpicture}
	\end{center}
	\caption{The systematic steps of learning with prior information.} \label{diagram_prior}
\end{figure*}

The system of learning with prior information is depicted in Figure \ref{diagram_prior}. The first step is to translate domain knowledge into prior information about the target hypothesis of $\mathcal{F}$, that is, to transform tacit knowledge about the problem at hand into properties that should be satisfied by $f^{\star}$ as consequence of it. For example, it is known that images are translation invariant, scale invariant, and depending on the domain are also rotation invariant. Indeed, if the learning task is to identify the presence of a given object in the image it should not matter if the object is centered or dislocated to the right (is translated), or if it occupies the majority of the image or is small relatively to its size (is scaled), as in any way the target classifier should indicate the presence of the object. Hence, one can learn on a hypotheses space $\mathcal{H}$ which contains only translation and scale invariant hypotheses, hoping to not lose the target hypothesis $f^{\star}$.

Once properties of $f^{\star}$ have been established, one can learn on a constrained hypotheses space $\mathcal{H}$, which contains all functions that satisfy such properties and, if prior information was right and $f^{\star}$ indeed satisfies the deduced properties, then there will be no approximation error. Fixed the hypotheses space, one learns a hypothesis in it from data.

\begin{figure*}[ht]
	\begin{center}
		\begin{tikzpicture}[scale=1, transform shape]
			\draw[fill = gray!20] (-7,-6) rectangle (1.75,-2);
			\draw[fill = bblue,rounded corners=1mm] (-4.75,-4) \irregularcircle{1.2cm}{1mm};
			
			\node at (-5.25,-2.65) {$\mathcal{H}$};
			\node at (-6.75,-1.8) {$\mathcal{F}$};
			
			\node at (-4.95,-4) {$f^{\star}$};
			\node (p1) at (-5.25,-4.25)[circle,fill,inner sep=1pt]{};
			
			\node at (-4.75,-4.75) {$\hat{h}$};
			\node (p2) at (-4.9,-4.65)[circle,fill,inner sep=1pt]{};
			
			\draw[-] (p1) -- (p2) node[pos=0.5,sloped,below] {$\Delta_{est}$};
		\end{tikzpicture}
	\end{center}
	\caption{Optimal insertion of prior information when $f^{\star} = h^{\star}$ and $\mathcal{H}$ has low complexity.} \label{model_prior}
\end{figure*}
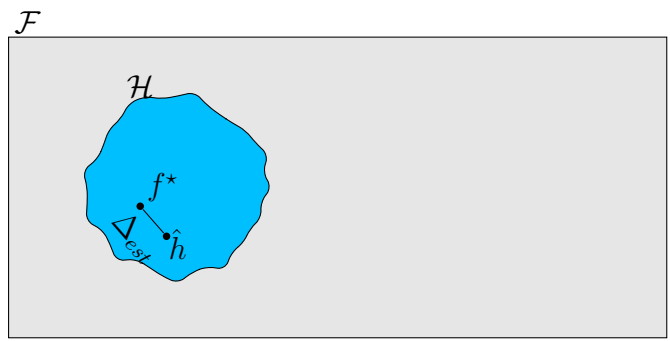

In the system of Figure \ref{diagram_prior} the trade-off between the generalization and approximation error of the hypotheses space is controlled constraining $\mathcal{F}$ by employing prior information about $f^{\star}$ so that it is either in $\mathcal{H}$, or is such that $L(f^{\star}) \approx L(h^{\star})$. The ideal learning scenario would be when prior information is right, so $f^{\star}$ is in the constrained hypotheses space $\mathcal{H}$, and $\mathcal{H}$ is \textit{simple} so with high probability, the generalization error with respect to $f^{\star}$ is small. This is the case illustrated in Figure \ref{model_prior}, in which $\mathcal{F}$ is quite complex, but one is learning on a simple hypotheses space $\mathcal{H}$ which contains the target hypothesis $f^{\star}$ of $\mathcal{F}$. Nevertheless, even when this is not the case, a proper insertion of prior information may cause the approximation error to decrease and, being $\mathcal{H}$ simple, one may have a small generalization error with high probability for a fixed sample size by \eqref{bound}.

Many domains which suffer from shortage of data have relied on a careful design of the hypotheses space. For instance, systems biology, a research area focused on the study of complex networks of molecular interactions existing in live organisms, is a field that to this day rely on a meticulous design of the hypotheses space to properly solve learning problems involving, for example, gene interactions, with very few samples of multidimensional data (see, for example, \cite{cubas2015linear,trepode2013pattern,barrera2007feature}). 

Not only shortage of samples, but also the lack of high computational power, has demanded the insertion of prior information. For instance, although the hand-made operators of Mathematical Morphology were useful to solve specific relatively simple problems, the need for manual design of operators limited their application in more complex tasks. However, in the 1990s there was a lack of computational power to perform a fully automatic design of operators, so at least a partial manual design was necessary to restrict the learning to a less complex hypotheses space where it could be performed with the data and computational resources available. Since this is an area on which the insertion of prior information has been the main focus since its inception, doing so to learn from data was a natural endeavor. See, for example, \cite{barrera1997automatic,barrera2000automatic,brun2003design,brun2004nonlinear,hirata1999design,dougherty2003hands,barrera2022mathematical}.

Although the use of prior information and domain knowledge was the usual approach in the early decades of Machine Learning, with the advent of cheap high computational power the area moved towards a fully automatic learning in detriment of a more careful design of the hypotheses space based on prior information. The efficiency of algorithms and the surprisingly good generalization of computing demanding black box learning frameworks led to a gold rush in the direction of a completely automated learning that requires little to none human intervention or prior information.

\section{High computational power in Machine Learning}
\label{SecHigh}

The high computational power of modern Machine Learning methods is usually directed to explicitly search for a hypotheses space, what is known in the literature as model selection, and to perform algorithm-oriented learning, in which the hypotheses space is disregarded, and the learning process is viewed as an algorithm. In this section, we discuss these usages of high computational power.

\subsection{Explicit search for a hypotheses space}

A first learning approach which employs high computational power is characterized by selecting a hypotheses space in a collection of candidates from data and then learning a hypothesis on it. This is a combinatorial problem, since requires a search on the set of candidate models, and is classically known in the literature as model selection \cite{massart2007concentration}. We use the word model as synonym of candidate hypotheses space.

Let $\mathbb{L} = \{\mathcal{H}_{1},\dots,\mathcal{H}_{n}\}$ be a collection of candidate models and fix a loss function $\widehat{L}: \mathbb{L} \to \mathbb{R}_{+}$ which associates to each candidate a \textit{loss} incurred when this hypotheses space is used for learning. The loss is calculated from sample $\mathcal{D}_{N}$, hence is dependent on data. Model selection means selecting a hypothesis space $\widehat{\mathcal{H}}$ in $\mathbb{L}$ which minimizes the loss:
\begin{linenomath}
	\begin{equation*}
		\widehat{L}(\widehat{\mathcal{H}}) \leq \widehat{L}(\mathcal{H}), \ \ \ \forall \mathcal{H} \in \mathbb{L}.
	\end{equation*}
\end{linenomath}
The loss function $\hat{L}$ is usually a penalization of the empirical risk of its minimizer $\hat{h}_{i}$:
\begin{linenomath}
	\begin{equation}
		\label{Pen_error}
		\widehat{L}(\mathcal{H}_{i}) = L_{\mathcal{D}_{N}}(\hat{h}_{i}) + \text{pen}(\mathcal{H}_{i})
	\end{equation}
\end{linenomath}
in which $\text{pen}: \mathbb{L} \to \mathbb{R}$ is a penalization function, that is increasing on the hypotheses space complexity. Other common losses estimators are obtained via a cross-validation or bootstrap procedure \cite{arlot2010survey,zhang1993model,zhang2015cross,kohavi1995,estimation}.

The minimization of \eqref{Pen_error} is an explicit attempt to control the trade-off between the complexity of the hypotheses space, related to its internal generalization error by \eqref{bound}, and the risk of its target hypothesis, i.e., its approximation error. The minimizer of \eqref{Pen_error} is such that the minimal empirical risk compensates the complexity, that is, a complex hypotheses space will be selected only if there is strong evidence that there exists a hypothesis in it with low expected risk. The performance of such control of the trade-off will be dependent on how well the choice of penalization can balance the hypotheses space complexity with its approximation error. There is a whole research area about penalization methods, and we refer to \cite{massart2007concentration} for an overview.

In order to compute $\widehat{\mathcal{H}}$ it is in principle necessary to perform an exhaustive search of $\mathbb{L}$, which is computationally demanding, and actually unfeasible when the cardinality of $\mathbb{L}$ is too great, as is often the case. Nevertheless, the problem of model selection is usually solved heuristically, applying algorithms that do not necessarily compute $\widehat{\mathcal{H}}$, but that return efficiently a suboptimal model.

Model selection substitutes the insertion of prior information to constraint $\mathcal{F}$ with the automatic selection of a hypotheses space through a computer demanding procedure. A classical example is variable selection or dimensionality reduction, in which the input $X$ has many variables, and the candidate models represent the hypotheses which depend solely on variables in a given subset \cite{guyon2003,john1994}. Model selection in this instance means selecting a subset of variables hopping that $f^{\star}$ depends solely on variables in the subset, so it is in the selected hypotheses space. This problem seeks to solve what is known as the curse of dimensionality that prevents the proper learning when the number of variables in $X$ is too great \cite{raudys1980dimensionality,zollanvari2020theoretical}. There are many efficient algorithms that solve heuristically the variable selection problem \cite{MG63,SS93,Whi71,PNK94,SPN99,SS89,NF77,u-curve3,ucurveParallel,u-curve2,featsel,reis2018,u-curve1}.

Although model selection substitutes the careful design of the hypotheses space from prior information, domain knowledge may be employed to choose the candidate models in a way which reflect some known properties of the target hypothesis. In this case, prior information is inserted into the set of candidate hypotheses spaces, as will be discussed in more details in Section \ref{Sec_comb}.

\subsection{Algorithm-oriented Machine Learning}

In any learning task, modern or otherwise, there is a hypotheses space and a learning algorithm, which processes a sample and returns a hypothesis. However, the hypotheses space may be implicit from the chosen algorithm and may not be given any thought by the practitioner, in a way that the design of learning actually reduces to the choice of an algorithm. With the dissemination of high computational power, modern Machine Learning applications are mainly algorithm-oriented and the hypotheses space became a merely theoretical concept.

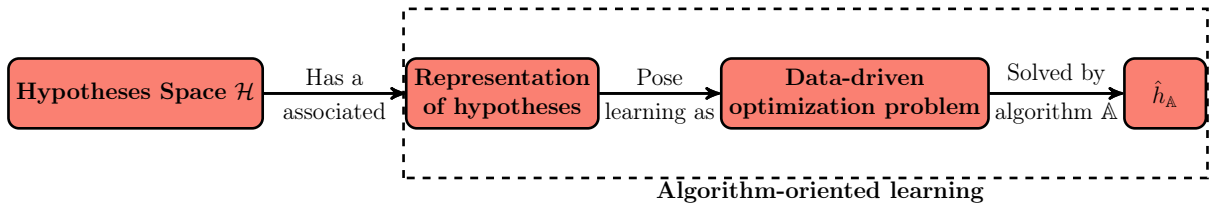
\begin{figure*}[ht]
	\begin{center}
		\begin{tikzpicture}[scale=0.75, transform shape]
			\tikzstyle{hs} = [rectangle,draw=black,fill = color2bg,rounded corners, minimum height=3em, minimum width=3.5em, node distance=2.5cm, line width=1pt,align=center]
			\tikzstyle{ns} = [rectangle,draw=white,rounded corners, minimum height=3em, minimum width=3.5em, node distance=2.5cm, line width=1pt]
			
			\draw[dashed,line width=1pt] (-7.25,-1.5) rectangle (7,1.5);
			
			\node[hs] (a1) at (-12,0) {\textbf{Hypotheses Space} $\mathcal{H}$};
			\node[hs] (a2) at (-5.5,0) {\textbf{Representation}\\ \textbf{of hypotheses}};
			\node[hs] (a3) at (0.75,0) {\textbf{Data-driven}\\ \textbf{optimization problem}};
			\node[hs] (a4) at (6.25,0) {$\hat{h}_{\mathbb{A}}$};
			\node (tex) at (0.1375,-1.75) {\textbf{Algorithm-oriented learning}};
			
			\begin{scope}[line width=1pt]
				\draw[->] (a1) -- (a2) node[pos=0.5,sloped,above] {Has a};
				\draw[->] (a1) -- (a2) node[pos=0.5,sloped,below] {associated};
				\draw[->] (a2) -- (a3) node[pos=0.5,sloped,above] {Pose};
				\draw[->] (a2) -- (a3) node[pos=0.5,sloped,below] {learning as};
				\draw[->] (a3) -- (a4) node[pos=0.5,sloped,above] {Solved by};
				\draw[->] (a3) -- (a4) node[pos=0.5,sloped,below] {algorithm $\mathbb{A}$};
			\end{scope}
		\end{tikzpicture}
	\end{center}
	\caption{General steps of learning from data. Once the hypotheses space $\mathcal{H}$ is fixed, a representation for its hypotheses is chosen, and a data-driven optimization problem is posed and then solved to learn hypotheses. In algorithm-oriented learning there is no explicit definition of a hypotheses space and the learning starts with the choice of a representation for the classifiers which, although generates a hypotheses space, the properties of its hypotheses are not known or cannot be directly identified in the representation parameters.} \label{break_learn}
\end{figure*}

The classic steps to solve a learning problem are presented in Figure \ref{break_learn}. First, one has to choose a hypotheses space $\mathcal{H}$ and, as discussed in Section \ref{SecPrior}, prior information and domain knowledge should guide such a choice. Then, one sets an efficient way to represent the hypotheses in $\mathcal{H}$ by a finite number of parameters, so the learning problem can be represented as a data-driven optimization problem whose solution is the parameters which optimize some quantity of interest associated to the classification of instances in a dataset. Finally, the posed optimization problem is solved by an algorithm $\mathbb{A}$ and its solution are the parameters which represent a hypothesis $\hat{h}_{\mathbb{A}}$, that by design is in $\mathcal{H}$.

When high computational power is not available, the data-driven optimization problem and the algorithm $\mathbb{A}$ should be efficient, for otherwise it may be difficult to compute $\hat{h}_{\mathbb{A}}$. In order to make them more efficient, one has to simplify the hypotheses space $\mathcal{H}$, more efficiently represent the hypotheses in it and pose the learning problem as a more efficient optimization problem. These manners of improving the efficiency of learning have indeed been the main topics in Machine Learning research when computer resources were scarce \cite{kotsiantis2006machine}.

Take for instance Mathematical Morphology operators and linear classifiers. The main topic in Mathematical Morphology research is how to represent operators with given properties \cite{banon1991minimal,barrera2001sup,barrera2022mathematical}, through minimal representations and combinations of simple building blocks. The linear classifiers, or perceptrons \cite{rosenblatt1957}, the first proposed classifiers, form one of the most simple hypotheses spaces, whose hypotheses are represented by the parameters of hyperplanes. Furthermore, they can be learned via an efficient optimization problem, such as that posed by linear support vector machines \cite{boser1992training}.

With the advent of high computational power, more complex optimization problems can now be solved, and the hypotheses can have more complex representations, what led to algorithm-oriented learning. As illustrated in Figure \ref{break_learn}, in algorithm-oriented learning one come up with a form of representing classifiers and pose a data-driven optimization problem whose solution are parameters which fully characterize a classifier $\hat{h}_{\mathbb{A}}$. In this framework, there is not an explicit definition of a hypotheses space.

Actually, it may not even be known to the practitioner what hypotheses can be represented by the chosen representation, so the hypotheses space generated by it, although well defined mathematically\footnote{A given representation of hypotheses generates the hypotheses space composed by all hypotheses that can be represented by it.}, is not exactly known. In special, the properties of the hypotheses generated by the representation, and even the properties of the learned hypothesis $\hat{h}_{\mathbb{A}}$, may be unknown. This means that, even though the learned classifier may generalize very well, that is, have low risk when classifying instances not in the dataset used to learn it, it is not interpretable and the practitioner do not really know what it does. We call ``black box'' the classifiers learned by algorithm oriented learning approaches that the practitioner do not know properties. 

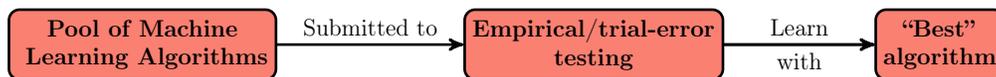
\begin{figure}[ht]
	\begin{center}
		\begin{tikzpicture}[scale=0.75, transform shape]
			\tikzstyle{hs} = [rectangle,draw=black,fill = color2bg,rounded corners, minimum height=3em, minimum width=3.5em, node distance=2.5cm, line width=1pt,align=center]
			\tikzstyle{ns} = [rectangle,draw=white,rounded corners, minimum height=3em, minimum width=3.5em, node distance=2.5cm, line width=1pt]
			
			\centering
			\node[hs] (b1) at (-14,-4) {\textbf{Pool of Machine}\\ \textbf{ Learning Algorithms}};
			\node[hs] (b2) at (-6,-4) {\textbf{Empirical/trial-error}\\ \textbf{testing}};
			\node[hs] (b3) at (0.1375,-4) {\textbf{``Best''}\\ \textbf{algorithm}};
			
			\begin{scope}[line width=1pt]
				\draw[->] (b1) -- (b2) node[pos=0.5,sloped,above] {Submitted to};
				\draw[->] (b2) -- (b3) node[pos=0.5,sloped,above] {Learn};
				\draw[->] (b2) -- (b3) node[pos=0.5,sloped,below] {with};
			\end{scope}
		\end{tikzpicture}
	\end{center}	
	\caption{Steps to choose the algorithm in algorithm-oriented Machine Learning.} \label{NEW}
\end{figure}

Instead of fixing or choosing a hypotheses space, in algorithm-oriented Machine Learning one has to choose an algorithm to learn from data, which is understood here as composed by a representation of classifiers, a data-driven optimization problem and an actual algorithm that computes $\hat{h}_{\mathbb{A}}$. This choice is usually performed as illustrated in Figure \ref{NEW}. There is a pool of learning algorithms implemented and widely available to which one resorts to solve a learning problem \cite{sarker2021machine,mahesh2020machine,gevorkyan2019review,purkait2019hands}. These algorithms are tested empirically in the available sample, and one of them is chosen based on its performance on the observed data. The ``best'' algorithm according to the tests is then employed to learn a classifier.

The pool of algorithms is actually stratified in algorithms that, according to empirical evidence and sometimes superficial domain knowledge, are suitable to solve each class of problems. For example, it is well established that convolutional neural networks (CNN) are the golden standard in image classification tasks \cite{albawi2017understanding,li2021survey}. Although it takes into account some domain knowledge, as the invariance to translation and scale of images, the features present in cutting-edge architectures were developed by trial and error, and empirical testing, rather than mathematically deduced. Therefore, when one is facing an image classification or object detection learning problem, he resorts to an empirically tuned CNN architecture and does not really need any domain knowledge or prior information to solve the learning problem.

In computing demanding learning algorithms, which are actually optimization algorithms, the control of the trade-off between the complexity of the classifier and its generalization quality is performed through constraints on the optimization problem in what is known as regularization \cite{bickel2006,micchelli2005,oneto2016} which are, in general, represented by a penalization of an empirical loss by the complexity of the hypothesis, usually measured by a norm of its parameters (see for example L1 and L2 regularization \cite{ng2004}).

Since the concept of hypotheses space plays no real role in algorithm-based learning, the trade-off is relative to the complexity and generalization quality of learned hypothesis $\hat{h}_{\mathbb{A}}$, rather than between the approximation error and complexity of the hypotheses space, and is controlled in the data-driven learning optimization problem rather than on the choice of hypotheses space.

The high performance of modern computer demanding learning methods evidences that high computational power may be sufficient for properly learning, as prior information was sufficient in classical methods, with a sample of fixed size, mitigating a possible lack of data. However, this high performance comes at the expense of the interpretability and knowledge about the properties of the learned classifier. A way to rescue the interpretability is to combine prior information with high computational power. This is the topic of the next section.

\section{Combining prior information with high computational power}
\label{Sec_comb}

Prior information is intrinsically related to the choice of the hypotheses space, which should contain hypotheses compatible with the perceived reality expressed by such information. Hence, in order to combine prior information with high computational power, one should insert constraints into the representation of hypotheses, and data-driven optimization problem which can be identified with constraints in the represented hypotheses space that are compatible with prior information. We will discuss the steps to achieve this goal when employing high computational power to explicitly search for a hypotheses space and to perform algorithm-oriented learning.

\subsection{Explicit search for a hypotheses space}
\label{Sec_SearchHS}

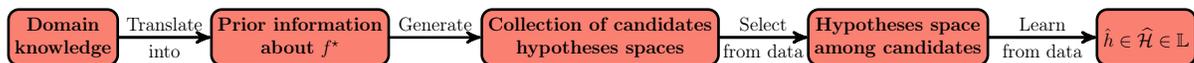
\begin{figure*}[ht]
	\begin{center}
		\begin{tikzpicture}[scale=0.6, transform shape]
			\tikzstyle{hs} = [rectangle,draw=black,fill = color2bg,rounded corners, minimum height=3em, minimum width=3.5em, node distance=2.5cm, line width=1pt,align=center]
			\tikzstyle{ns} = [rectangle,draw=white,rounded corners, minimum height=3em, minimum width=3.5em, node distance=2.5cm, line width=1pt]
			
			\node[hs] (a1) at (-13.5,0) {\textbf{Domain}\\ \textbf{knowledge}};
			\node[hs] (a2) at (-8.25,0) {\textbf{Prior information}\\ \textbf{about} $f^{\star}$};
			\node[hs] (a3) at (-1.6,0) {\textbf{Collection of candidates}\\ \textbf{hypotheses spaces}};
			\node[hs] (a4) at (5,0) {\textbf{Hypotheses space}\\ \textbf{among candidates}};
			\node[hs] (a5) at (10.5,0) {$\hat{h} \in \widehat{\mathcal{H}} \in \mathbb{L}$};
			
			\begin{scope}[line width=1pt]
				\draw[->] (a1) -- (a2) node[pos=0.5,sloped,above] {Translate};
				\draw[->] (a1) -- (a2) node[pos=0.5,sloped,below] {into};
				\draw[->] (a2) -- (a3) node[pos=0.5,sloped,above] {Generate};
				\draw[->] (a3) -- (a4) node[pos=0.5,sloped,above] {Select};
				\draw[->] (a3) -- (a4) node[pos=0.5,sloped,below] {from data};
				\draw[->] (a4) -- (a5) node[pos=0.5,sloped,above] {Learn};
				\draw[->] (a4) -- (a5) node[pos=0.5,sloped,below] {from data};
			\end{scope}
		\end{tikzpicture}
	\end{center}
	\caption{The systematic steps of model selection when prior information is combined with high computational power to select a hypotheses space from data.} \label{diagram_prior2}
\end{figure*}

When selecting the hypotheses space from data, prior information should be inserted into the collection of candidate hypotheses spaces, as illustrated in Figure \ref{diagram_prior2}. In this instance, domain knowledge is translated into prior information about $f^{\star}$ that generate a collection of candidate hypotheses spaces from which one is selected based on data and a hypothesis is learned on it. This approach is preferable when prior information is not strong enough to generate a single hypotheses space, but can be taken advantage of to select a hypotheses space from data.

An example of this approach is variable selection, when there is prior information that some variables are spurious and the target $h^{\star}$ of a fixed hypotheses space $\mathcal{H}$ with finite VC dimension does not depend on all of them. In this case, it is not known which variables the target depends on and this prior information can be better employed to obtain a collection of candidate hypotheses spaces rather than delimit just one. In this approach, first of all, if $\mathcal{F}$ has infinite VC dimension one needs to delimit the learning to a hypotheses space $\mathcal{H}$ with finite VC dimension. Then, with the information that $h^{\star}$ does not depend on all variables, one consider as candidate models the sets of functions in $\mathcal{H}$ which depend solely on variables in each subset of variables. Observe that $\mathcal{H}$ is a candidate hypotheses space, since it is formed by the functions which depend on variables in the whole set of variables.

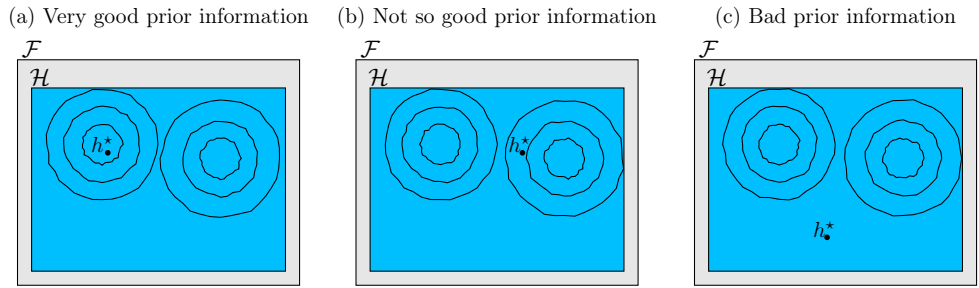
\begin{figure*}[ht]
	\begin{center}
		\begin{tikzpicture}[scale=0.75, transform shape]
			\draw[fill = gray!20] (-7,-6) rectangle (-2,-2);
			\draw[fill = bblue] (-6.75,-5.75) rectangle (-2.25,-2.5);
			
			\draw[fill = gray!20] (-1,-6) rectangle (4,-2);
			\draw[fill = bblue] (-0.75,-5.75) rectangle (3.75,-2.5);
			
			\draw[fill = gray!20] (5,-6) rectangle (10,-2);
			\draw[fill = bblue] (5.25,-5.75) rectangle (9.75,-2.5);
			
			\draw (-5.5,-3.5) \irregularcircle{10pt}{0.75pt};
			\draw (-5.5,-3.5) \irregularcircle{19pt}{0.75pt};
			\draw (-5.5,-3.5) \irregularcircle{28pt}{0.75pt};
			\node at (-5.5,-3.5) {\small$h^{\star}$};
			\node at (-5.4,-3.65)[circle,fill,inner sep=1pt]{};
			
			\draw (-3.4,-3.75) \irregularcircle{10pt}{0.75pt};
			\draw (-3.4,-3.75) \irregularcircle{19pt}{0.75pt};
			\draw (-3.4,-3.75) \irregularcircle{30pt}{0.75pt};

			\node at (-6.6,-2.3) {$\mathcal{H}$};
			\node at (-6.75,-1.8) {$\mathcal{F}$};
			\node at (-4.5,-1.25) {\small (a) Very good prior information};
			
			\draw (0.5,-3.5) \irregularcircle{10pt}{0.75pt};
			\draw (0.5,-3.5) \irregularcircle{19pt}{0.75pt};
			\draw (0.5,-3.5) \irregularcircle{28pt}{0.75pt};
			\node at (1.9,-3.5) {\small$h^{\star}$};
			\node at (1.95,-3.65)[circle,fill,inner sep=1pt]{};
			
			\draw (2.7,-3.75) \irregularcircle{10pt}{0.75pt};
			\draw (2.7,-3.75) \irregularcircle{19pt}{0.75pt};
			\draw (2.7,-3.75) \irregularcircle{29.5pt}{0.75pt};
			
			\node at (1.5,-1.25) {\small (b) Not so good prior information};
			\node at (-0.6,-2.3) {$\mathcal{H}$};
			\node at (-0.75,-1.8) {$\mathcal{F}$};
			
			\draw (6.5,-3.5) \irregularcircle{10pt}{0.75pt};
			\draw (6.5,-3.5) \irregularcircle{19pt}{0.75pt};
			\draw (6.5,-3.5) \irregularcircle{28pt}{0.75pt};
			\node at (7.3,-5) {\small$h^{\star}$};
			\node at (7.35,-5.15)[circle,fill,inner sep=1pt]{};
			
			\draw (8.7,-3.75) \irregularcircle{10pt}{0.75pt};
			\draw (8.7,-3.75) \irregularcircle{19pt}{0.75pt};
			\draw (8.7,-3.75) \irregularcircle{29.5pt}{0.75pt};
			
			\node at (7.5,-1.25) {\small (c) Bad prior information};
			\node at (5.4,-2.3) {$\mathcal{H}$};
			\node at (5.25,-1.8) {$\mathcal{F}$};

		\end{tikzpicture}
	\end{center}
	\caption{Three examples of inserting prior information to choose a collection of candidate models. A hypotheses space $\mathcal{H}$ with finite VC dimension is chosen and, based on prior information, a collection of candidate models, represented by the circles, whose complexity is proportional to their size, is fixed. We assume that the whole hypotheses space $\mathcal{H}$ is also a candidate. (a) Prior information is very good since the target hypothesis is within a candidate model with low complexity, hence, if this hypotheses space is selected from data, the complexity-approximation error trade-off is controlled since there is no bias relative to learning on the whole $\mathcal{H}$ and the complexity is the least among the unbiased candidates. (b) Prior information is not so good, since, although $h^{\star}$ is in a candidate hypotheses space less complex than $\mathcal{H}$, it is among the most complex. (c) Prior information is bad, since the target hypothesis is not within a simpler candidate hypotheses space, and can only be learned if the whole $\mathcal{H}$ is selected from data.} \label{candidates}
\end{figure*}

If prior information is right, the learning is within the scenarios depicted in Figure \ref{candidates} (a) and (b), which represents a hypotheses space $\mathcal{H}$ with finite VC dimension and a collection of candidates, some of which form a chain $\mathcal{H}_{1} \subset \dots \subset \mathcal{H}_{n}$. In (a) and (b), the target is within a hypotheses space apart from $\mathcal{H}$ hence, if it is selected from data, the learning will be performed without a bias relative to $\mathcal{H}$. Furthermore, if the scenario is actually that of (a) then, if the most simple hypotheses space containing $h^{\star}$ is selected, the learning would be not only without bias relative to $\mathcal{H}$, but also performed on the least complex hypotheses space with this property among the candidates, implying a control of the complexity-approximation error trade-off. On the other hand, if the scenario is that of (b), in order to learn without a bias relative to $\mathcal{H}$ one would have to learn on a hypotheses space within the more complex among the candidates, but that is still simpler than $\mathcal{H}$ and a lesser generalization error may be achieved.

However, prior information may be wrong, and the target hypothesis depends on all variables. In this case, the learning scenario would be that represented in Figure \ref{candidates} (c) and the learning would be without bias relative to $\mathcal{H}$ only if the whole space was selected to learn on, but in this case, there would be no control of the complexity in the trade-off.

The example discussed above and the depiction of Figure \ref{candidates} illustrate what it means for a collection of candidate models to be compatible with the available prior information. It is compatible when the target hypothesis is within a candidate model with a complexity lesser than that of $\mathcal{H}$, and the degree of compatibility is represented by the least complexity of a candidate model that contains the target hypothesis. 

Observe there is no guarantee that a hypotheses space containing $h^{\star}$ will be selected, but if the collection of candidate models is compatible with prior information then there is a chance the learning will be performed on a hypotheses space simpler than $\mathcal{H}$ without any bias relative to it. In a perfect scenario, such as that of Figure \ref{candidates} (a), if computational power is high enough to properly search for a hypotheses space and the simplest hypotheses space containing $h^{\star}$ is selected, then prior information is optimally combined with high computational power to control the complexity-approximation trade-off: the approximation error is minimum within $\mathcal{H}$, i.e., there is no bias, and the complexity of the hypotheses space is the lowest among the unbiased candidates. 

\subsection{Algorithm-oriented Machine Learning}

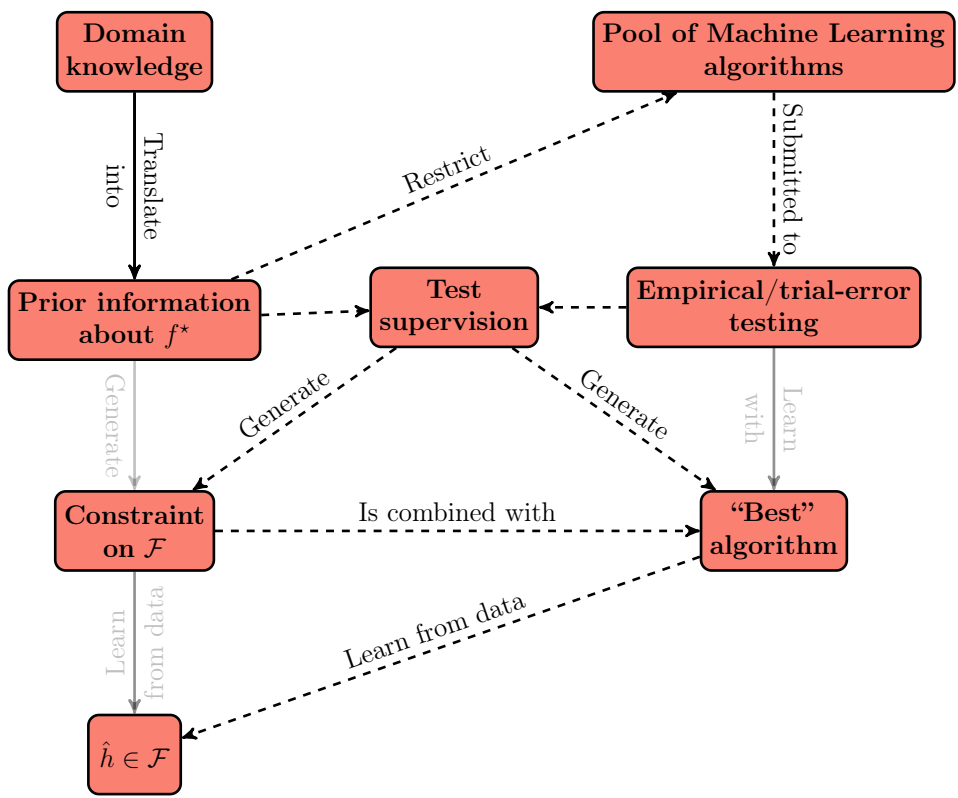
\begin{figure*}[ht]
	\begin{center}
		\begin{tikzpicture}[scale=0.85, transform shape]
			\tikzstyle{hs} = [rectangle,draw=black,fill = color2bg,rounded corners, minimum height=3em, minimum width=3.5em, node distance=2.5cm, line width=1pt,align=center]
			\tikzstyle{ns} = [rectangle,draw=white,rounded corners, minimum height=3em, minimum width=3.5em, node distance=2.5cm, line width=1pt]
			
			\node[hs] (a1) at (5,0.5) {\textbf{Pool of Machine Learning}\\ \textbf{algorithms}};
			\node[hs] (a2) at (5,-3.5) {\textbf{Empirical/trial-error}\\ \textbf{testing}};
			\node[hs] (a3) at (5,-7) {\textbf{``Best''}\\ \textbf{algorithm}};
			\node[hs] (a4) at (0,-3.5) {\textbf{Test}\\ \textbf{supervision}};
			
			\node[hs] (b1) at (-5,0.5) {\textbf{Domain}\\ \textbf{knowledge}};
			\node[hs] (b2) at (-5,-3.7) {\textbf{Prior information}\\ \textbf{about} $f^{\star}$};
			\node[hs] (b3) at (-5,-7) {\textbf{Constraint}\\ \textbf{on} $\mathcal{F}$};
			\node[hs] (b4) at (-5,-10.5) {$\hat{h} \in \mathcal{F}$};

			\begin{scope}[line width=1pt]
				\draw[->,dashed] (a1) -- (a2) node[pos=0.5,sloped,above] {Submitted to};
				\draw[->,opacity=0.25] (a2) -- (a3) node[pos=0.5,sloped,above] {Learn};
				\draw[->,opacity=0.25] (a2) -- (a3) node[pos=0.5,sloped,below] {with};
				
				\draw[->] (b1) -- (b2) node[pos=0.5,sloped,above] {Translate};
				\draw[->] (b1) -- (b2) node[pos=0.5,sloped,below] {into};
				\draw[->,opacity=0.25] (b2) -- (b3) node[pos=0.5,sloped,below] {Generate};
				\draw[->,opacity=0.25] (b3) -- (b4) node[pos=0.5,sloped,above] {Learn};
				\draw[->,opacity=0.25] (b3) -- (b4) node[pos=0.5,sloped,below] {from data};
				
				\draw[->,dashed] (b2) -- (a1) node[pos=0.5,sloped,above] {Restrict};
				\draw[->,dashed] (b2) -- (a4);
				\draw[->,dashed] (a2) -- (a4);
				\draw[->,dashed] (a4) -- (b3) node[pos=0.5,sloped,above] {Generate};
				\draw[->,dashed] (a4) -- (a3) node[pos=0.5,sloped,above] {Generate};
				\draw[->,dashed] (b3) -- (a3) node[pos=0.5,sloped,above] {Is combined with};
				\draw[->,dashed] (a3) -- (b4) node[pos=0.5,sloped,above] {Learn from data};	
			\end{scope}
		\end{tikzpicture}
	\end{center}
	\caption{Combining prior information with high-performance Machine Learning algorithms.} \label{diagram_combine}
\end{figure*}

The insertion of prior information in algorithm-oriented Machine Learning can be performed by merging the steps of learning with prior information and algorithm-oriented Machine Learning presented in Figures \ref{diagram_prior} and \ref{NEW}, respectively, as illustrated in Figure \ref{diagram_combine}. In this approach, domain knowledge should be translated into prior information about $f^{\star}$ implying a restriction on the pool of Machine Learning algorithms that should be compatible with the believed properties of $f^{\star}$. The restricted pool of Machine Learning algorithms is submitted to trial-error testing, but this test should be supervised, so the learned hypothesis has at least some of the believed properties of $f^{\star}$. From testing follows the best algorithm and a constraint on $\mathcal{F}$ that generates a hypotheses space $\mathcal{H}$ containing only the hypotheses representable by the best algorithm. The constrained hypotheses space is combined with the best algorithm to learn a hypothesis.

The main difficulties of this approach are related to the lack of understanding about the algebraic properties of the parameters representing the hypotheses, so one cannot constrain $\mathcal{F}$ in a desired manner by constraining the parameters. In order to better understand what a fully understanding of the algebraic properties of parameters would be, we present the example of learning crescent real-valued functions.

Let $\mathcal{F} = \{f: \mathbb{R} \to \mathbb{R}\}$ be the set of continuous real-valued functions and consider the square mean risk
\begin{equation*}
	L(f) = E\left[\left(f(X) - Y\right)^{2}\right]
\end{equation*}
as the risk of each $f \in \mathcal{F}$. We assume there is prior information about $f^{\star}$ implying that it is a crescent function for $x \geq 0 $ and that $f^{\star}(x) > 0$ for all $x \in \mathbb{R}$. Since the VC dimension of $\mathcal{F}$ is infinite, we should restrict the learning to a subset of it, and we will choose as $\mathcal{H} \subset \mathcal{F}$ the degree two polynomials, that is,
\begin{equation*}
	\mathcal{H} = \{h(x) = a_{0} + a_{1} x + a_{2} x^{2}: a_{0}, a_{1}, a_{2} \in \mathbb{R}\}.
\end{equation*}

Observe that the properties of $f^{\star}$ can be inferred from $h$ and its derivative $h^{\prime}$, as crescent functions have a positive derivative, and we can consider only hypotheses satisfying
\begin{align}
	\label{ineq1}
	&h^{\prime}(x) = a_{1} + 2 a_{2} x > 0, \forall x \geq 0  \implies \frac{a_{1}}{2 a_{2}} > - x, \forall x \geq 0 \implies \frac{a_{1}}{2 a_{2}} > 0 \\ \label{ineq2}
	&h(x) = a_{0} + a_{1} x + a_{2} x^{2} > 0, \forall x \implies h\text{ has no real root } \implies  a_{1}^{2} - 4a_{0}a_{2} < 0.
\end{align}
Hence, the properties of $f^{\star}$ can be expressed as constraints in the parameters representing the hypotheses and one can consider as hypotheses space the degree two polynomials such that $a_{1}$ and $a_{2}$ have the same sign and $a_{1}^{2} - 4a_{0}a_{2} < 0$. In other words, the properties believed to be satisfied by $f^{\star}$ are identifiable in the parameters representing the hypotheses in $\mathcal{H}$.

\begin{figure*}[ht]
	\begin{center}
		\begin{tikzpicture}[scale=0.75, transform shape]
			\node at (-4.5,-1.5) {(a)};	
			\draw[fill = gray!20] (-7,-6) rectangle (-2,-2);
			
			\node at (1.5,-1.5) {(b)};	
			\draw[fill = gray!20] (-1,-6) rectangle (4,-2);
			\draw[fill = color2bg] (-0.5,-5.5) rectangle (3,-3);
			
			\node at (-5.5,-2.45) {$P_{1}$};
			\draw[fill=bblue,opacity=0.5] (-4.75,-3.5) \irregularcircle{30pt}{0.75pt};
			
			\node at (-2.8,-2.6) {$P_{2}$};
			\draw[fill=bblue2,opacity=0.5] (-3.25,-3.75) \irregularcircle{30pt}{0.75pt};
			\node at (-4,-3.5) {\small$f^{\star}$};
			\node at (-4,-3.6)[circle,fill,inner sep=1pt]{};
			
			\node at (-6.75,-1.8) {$\mathcal{F}$};
			
			\node at (0.5,-2.45) {$P_{1}$};
			\draw[fill=bblue,opacity=0.5] (1.25,-3.5) \irregularcircle{30pt}{0.75pt};
			
			\node at (3.2,-2.6) {$P_{2}$};
			\draw[fill=bblue2,opacity=0.5] (2.75,-3.75) \irregularcircle{30pt}{0.75pt};
			\node at (2,-3.5) {\small$f^{\star}$};
			\node at (2,-3.6)[circle,fill,inner sep=1pt]{};
			
			\node at (-0.8,-3) {$\Theta$};		
		\end{tikzpicture}
	\end{center}
	\caption{(a) The set of hypotheses that satisfy properties $P_{1}$ and $P_{2}$ when there is prior information that $f^{\star}$ satisfies both properties. (b) The parametric space $\Theta$ of all parameters which are possible solutions of a learning algorithm.} \label{HSprop}
\end{figure*}
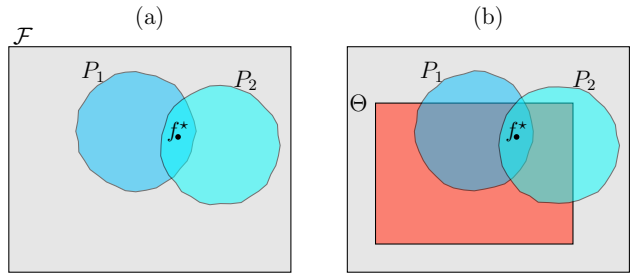

This case is illustrated in Figure \ref{HSprop}, in which $\mathcal{F}$ contains the sets of functions which are positive for all $x$ (property $P_{1}$) and crescent for $x \geq 0$ (property $P_{2}$), and $f^{\star}$ is in the intersection of these sets, represented in Figure \ref{HSprop} (a). In order to learn, we considered only functions that can be represented by degree two polynomials with parameters $a_{0}, a_{1}, a_{2}$ in a parametric space $\Theta = \mathbb{R}^{3}$, represented in Figure \ref{HSprop} (b). 

Although not all functions can be represented by degree two polynomials, some functions with properties $P_{1}$ and $P_{2}$ can be represented by them and, in this illustration, so can $f^{\star}$. In any case, even if $f^{\star}$ is not a polynomial, prior information can be inserted into the parameters representing the functions in $\mathcal{H}$ and the properties of $h \in \mathcal{H}$ can be inferred from its parameters. In special, properties $P_{1}$ and $P_{2}$ can be inferred by testing if inequalities $\eqref{ineq1}$ and $\eqref{ineq2}$ hold. 

In view of the framework in Figure \ref{diagram_combine}, the steps to insert prior information into algorithm-oriented Machine Learning in this simple synthetic example could be:
\begin{enumerate}
	\item Translate domain knowledge about the problem at hand into prior information about properties that $f^{\star}$ should satisfy. In this step, we concluded that $f^{\star}$ should satisfy properties $P_{1}$ and $P_{2}$ which are identified in its parameters by inequalities \eqref{ineq1} and \eqref{ineq2}.
	\item Restrict the pool of algorithms based on this prior information. Given the constraint generated by \eqref{ineq1} and \eqref{ineq2}, ordinary least squares cannot be applied to minimize the empirical mean square risk and other algorithms should be chosen to compute $\hat{h}$. In this pool of algorithms there may be some that in fact minimize the empirical risk under constraints \eqref{ineq1} and \eqref{ineq2}, and more efficient algorithms that in principle perform an unconstrained minimization of the empirical risk, but that may be supervised to perform the constrained optimization. For example, one could apply a gradient descent algorithm \cite{ruder2016overview}, that is an iterative algorithm which at each step update the parameters by going in the direction that most minimize the empirical risk locally, but supervise its trajectory by forbidding it to go into directions that violate \eqref{ineq1} or \eqref{ineq2}.
	\item Fixed a pool of learning algorithms, they are tested on empirical data. Some more efficient algorithms may not guarantee that constraints \eqref{ineq1} and \eqref{ineq2} are satisfied, even with supervision, but may return empirically good results, especially if prior information was not spot on. In this case, with a careful supervision of each algorithm, one that does not exactly guarantee the constraints may be selected as the best algorithm.
	\item The selection of the algorithm implies a constraint in $\mathcal{F}$ and generates a hypotheses space $\mathcal{H}^\prime$ containing only the hypotheses that can be learned by the algorithm. The best algorithm is then applied to minimize the empirical risk in $\mathcal{H}^\prime$ returning a hypothesis that may not satisfy \eqref{ineq1} and \eqref{ineq2} in case empirical evidence was strong against prior information, but that has properties which can be inferred from its parameters.	
\end{enumerate}

Learning with the steps described above has some advantages. On the one hand, some learning algorithms may be disregarded if they cannot represent hypotheses with the desired properties and, if prior information was right, then learning would be performed in an unbiased hypotheses space relative to $\mathcal{H}$ which is simpler than it, and hence better generalization could be achieved. On the other hand, if prior information was not completely right, considering algorithms that can also represent hypotheses that do not satisfy all constraints generated by prior information could actually improve generalization if empirical evidence points out to a hypothesis with better generalization that is incompatible with the mistaken prior information. Either way, the properties of the learned hypothesis are completely known, since can be inferred from its parameters.

The bottleneck of the steps illustrated in Figure \ref{diagram_combine} is the identification of the properties of the hypotheses in their parameters. If prior information cannot be translated into the parameters representing the hypotheses or into the optimization algorithm employed to learn, then the steps described above cannot be performed. Unfortunately, this is still the case of many modern Machine Learning methods, specially neural network, and a further study of this identification is a way to truly insert prior information into the learning process.

If this understanding of neural networks was achieved, then inserting prior information into the learning process could be as follows:
\begin{enumerate}
	\item Translate domain knowledge about the problem at hand into prior information about properties that $f^{\star}$ should satisfy and identify these properties with the parameters of a pool of neural networks architectures.
	\item Restrict the pool of architectures based on this prior information and consider only architectures that can represent hypotheses with the desired properties. Some architectures may represent only hypotheses with the properties, while others may represent these hypotheses and others.
	\item Fixed a pool of neural networks, test them on empirical data. To test architectures that also represent hypotheses without the desired properties, the test should be supervised to guarantee that it is not diverting too far away from prior information, but also to allow the learning of hypotheses that not fully satisfy prior information in case prior information was not completely right.
	\item The selection of an architecture implies a constraint in $\mathcal{F}$ and generates a hypotheses space $\mathcal{H}$ containing only the hypotheses that can be represented by the chosen architecture. The best architecture is then applied returning a hypothesis that may not fully satisfy prior information in case empirical evidence was strong against it, but that has properties which can be inferred from its parameters. Since these properties are known, the returned hypothesis is at least not a fully black box.
\end{enumerate}

The steps described above actually represent the search for a neural network architecture. There is a whole area of research, called neural architecture search \cite{elsken2019}, concerned with this problem. However, since the properties of a hypothesis generated by an architecture cannot be completely identified in its parameters, this search is usually performed by employing even more computational power to build an architecture from data, rather than inserting prior information to generate a collection of candidate architectures which are then searched, in a framework analogous to that discussed in Section \ref{Sec_SearchHS}.

The advantages of selecting architectures by the steps above are significant, since it would transform the development of neural network architectures from an empirical to a theoretical endeavor. This would probably imply an increasing performance of neural networks, specially in domains where they do not perform very well nowadays, since architectures which represent hypotheses compatible with prior information of each domain could be surgically chosen rather than being found by trial and error guided by empirical evidence. 

Furthermore, there would be a gain in interpretability since the whole process of selection and training of an architecture would be supervised and guided by prior information in a way that \textit{posterior} information about the learned hypothesis is readily available. Such a framework would benefit from the qualities of both learning with prior information, such as interpretability and oversight of the hypotheses space and learning procedure, and of learning with high computational power, such as computational efficiency and high generalization performance. Therefore, merging prior information with computation demanding methods is a promising way of increasing the performance and understanding of Machine Learning methods.

\section{Final Remarks}
\label{SecFR}

Both prior information and high computational power play important roles in machine learning, but approaches based on them have distinct levels of interpretability and performance. While prior information is vital to employ machine learning as an inductive inference tool, high computational power is vital for employing it as an instrument to solve practical problems. Both approaches have qualities which could be better taken advantage of by combining them, seeking to maximize the understanding and performance of machine learning methods.

Although this combination has been performed in the literature, specially under a Bayesian perspective \cite{theodoridis2015machine,tipping2003bayesian,snoek2012practical}, modern learning algorithms are, in general, black boxes and prior information is not strongly employed. A path to combine prior information with high computational power passes through the identification of properties of a classifier in the parameters representing them. In the context of neural networks, this means identifying properties of the classifiers in the architecture and parameters of a neural network. This would imply a better understanding of the estimated classifier and could aid on the design and training of neural networks, so a theoretical study of this identification is promising.

However, this research would be basic and theoretical, and would not immediately bring an enhancement to the performance of state-of-the-art machine learning methods. In an ever more instrumentalist research area, there is not much interest in research on basic questions and open problems pertaining to some theoretical aspects of machine learning, such as the identification of properties in parameters, have not been fully studied.

Even though studying these problems do not enhance the performance of modern methods right away, it may subsidize future empirical researches and enhance not only the understanding of the methods, but also their performance, as discussed in Section \ref{Sec_comb}. Therefore, we expect that this work may arouse a discussion about not only the role of prior information and computational power in inductive and instrumentalist machine learning, but also point out the need of basic theoretical research in this area.

\section*{Acknowledgments}

D. Marcondes was funded by grant \#22/06211-2, São Paulo Research Foundation (FAPESP). J. Barrera was funded by grants \#14/50937-1 and \#2020/06950-4, São Paulo Research Foundation (FAPESP).

\bibliographystyle{plain}
\bibliography{ref}

\end{document}